\documentclass[runningheads]{llncs}

\usepackage{eccv}

\usepackage{eccvabbrv}

\usepackage{graphicx}
\usepackage{booktabs}
\usepackage{dsfont}

\usepackage[accsupp]{axessibility}  

\usepackage{hyperref}

\usepackage{orcidlink}

\begin{document}

\title{Multi-view Pose Fusion for Occlusion-Aware 3D Human Pose Estimation} 


\author{Laura Bragagnolo\orcidlink{0009-0007-8096-4588} \and
Matteo Terreran\orcidlink{0000-0001-9862-8469} \and
Davide Allegro\orcidlink{0009-0008-1180-9290} \and
Stefano Ghidoni\orcidlink{0000-0003-3406-8719}}

\authorrunning{L. Bragagnolo et al.}

\institute{Intelligent Autonomous Systems Lab, Department of
Information Engineering, University of Padova, 35131 Padua, Italy \\
\email{\{bragagnolo, terreran, allegrodav, ghidoni\}@dei.unipd.it}}

\maketitle

\begin{abstract}
    Robust 3D human pose estimation is crucial to ensure safe and effective human-robot collaboration. Accurate human perception, however, is particularly challenging in these scenarios due to strong occlusions and limited camera viewpoints. Current 3D human pose estimation approaches are rather vulnerable in such conditions.
    In this work we present a novel approach for robust 3D human pose estimation in the context of human-robot collaboration.
    Instead of relying on noisy 2D features triangulation, we perform multi-view fusion on 3D skeletons provided by absolute monocular methods. Accurate 3D pose estimation is then obtained via reprojection error optimization, introducing limbs length symmetry constraints.
    We evaluate our approach on the public dataset Human3.6M and on a novel version Human3.6M-Occluded, derived adding synthetic occlusions on the camera views with the purpose of testing pose estimation algorithms under severe occlusions. We further validate our method on real human-robot collaboration workcells, in which we strongly surpass current 3D human pose estimation methods.
    Our approach outperforms state-of-the-art multi-view human pose estimation techniques and demonstrates superior capabilities in handling challenging scenarios with strong occlusions, representing a reliable and effective solution for real human-robot collaboration setups.
\end{abstract}

\section{Introduction}
\label{sec:intro}

Three-dimensional human pose estimation is a fundamental task in computer vision, relevant for many applications, such as virtual reality~\cite{lin2010augmented}, sport analysis~\cite{bridgeman2019multi} and human action recognition~\cite{peng2023delving}.  
Accurate 3D human pose estimation is crucial for human-robot collaboration. Reliable 3D human perception is essential for robots to understand the exact position and motion of humans, effectively preventing collisions and ensuring safety. Human pose monitoring also helps in assessing and improving worker ergonomics, enhancing overall workplace health and productivity. Moreover, robust 3D human pose estimation is critical for precise and prompt recognition of human actions and intentions, that can enable smooth and efficient human-robot collaboration~\cite{terreran2022skeleton,terreran2023general}. 

In industrial scenarios, accurate human pose estimation is a real challenge. The typical setting includes a multi-camera setup that frames the human-robot collaboration area---cameras are usually very tilted and often not synchronized. Moreover, the human body is often only partially visible, as surrounding obstacles and equipment often occlude the subject.
Such conditions are very under-represented in 3D human pose estimation datasets, which are usually collected in very controlled environments.

As multiple viewpoints are often available in human-robot collaboration workcells, most commonly adopted solutions are multi-view human pose estimation methods, that can accurately estimate 3D human joint position when camera calibration parameters are known. Most state-of-the-art multi-view approaches are, however, based on triangulation of 2D keypoints~\cite{iskakov2019learnable, chen2022structural, zhang2021adafuse}, so they cannot handle significant amounts of occlusions and clutter, as 2D pose estimation quality quickly degrades under these conditions and good predictions cannot be recovered.
Recent monocular 3D human pose estimation methods have achieved remarkable results on public benchmarks~\cite{gong2023diffpose, zhang2022mixste} and exhibit strong generalization to new and diverse scenarios~\cite{joo2021exemplar, zhang2020object}. On the other hand, when huge parts of the body are not visible, they tend to struggle with precise absolute scale prediction and reliable 3D joint configuration.
For these reasons, current human pose estimation systems are unsuitable for real industrial scenarios~\cite{boldo2024real}.

In this work, we propose a novel multi-view 3D human pose estimation method based on the fusion of 3D keypoints from multiple viewpoints to achieve robust and reliable human pose prediction in human-robot collaboration settings, characterized by the presence of heavy occlusions and few camera viewpoints.
While many current approaches are based on multi-view fusion of 2D features and significantly suffer when human joints are obscured, we take as input 3D human pose predictions given by absolute monocular methods.
To mitigate joints position artifacts produced in presence of large occlusions, we leverage multiple viewpoints and introduce a novel multi-view 3D keypoint fusion module to merge single-view 3D pose predictions.
Considering per-joint reprojection errors on all camera views, we can effectively correct body pose hallucinations and scale recovery errors. Accurate 3D pose localization is then achieved thanks to an optimization procedure on the fused pose, considering reprojection errors on 2D views. The final prediction is further refined considering body symmetry constraints during optimization. 
With respect to previous work~\cite{guidolin2023hi, chen2022structural} we do not require any prior information or statistics about human limbs length.

We validate our method on data collected from real human-robot collaboration workcells and on Human3.6M-Occluded, a novel benchmark derived from the public dataset Human3.6M~\cite{ionescu2013human3} adding synthetic occlusions on camera views.

To summarise, the main contributions of this paper are:
\begin{enumerate}
    \item a multi-view 3D human pose estimation method based on per-joint multi-view fusion and optimization, designed to tackle common problems in complex real scenarios, such as occlusions;
    \item limbs length symmetry constraints to refine 3D human pose, that do not depend on any prior information or statistics;
	\item an analysis on robustness and generalization capabilities of multi-view 3D human pose estimation methods in real human-robot collaboration workcells;
    \item the novel benchmark Human3.6M-Occluded, to test pose estimation algorithms under severe occlusions;
    \item extensive evaluation and ablation studies considering multi-view 3D human pose estimation methods on the public dataset Human3.6M and on Human3.6M-Occluded.    
\end{enumerate}

\section{Related Work}
\label{sec:related-work}

\subsubsection{Multi-view 3D Human Pose Estimation}
Multi-view human pose estimation methods leverage multiple camera viewpoints to provide accurate human pose predictions in 3D space. They typically include two steps: first they extract 2D features from each view independently, and then reconstruct 3D information via triangulation~\cite{iskakov2019learnable, zhang2021adafuse, he2020epipolar}. If triangulation-based approaches are very precise when provided with good quality 2D predictions, occluded environments are traditionally challenging for these methods, as 2D estimated poses tend to be very noisy.
Some approaches try to refine 2D predictions and triangulation incorporating epipolar geometry constraints: the epipolar transformer~\cite{he2020epipolar} learns attention weights to fuse pixels along the epipolar line in neighboring views; Ma \etal~\cite{ma2021transfusion} introduce Transfusion, a transformer framework for multi-view 3D pose estimation, aiming at directly improving 2D predictions integrating information from different views via geometry position encoding. Similarly,~\cite{moliner2024geometry, wan2023view} propose to leverage geometric correlations between views to refine 2D pose estimations.
Zhang \etal~\cite{zhang2021adafuse} present AdaFuse, which tackles human pose estimation in presence of occlusions by learning an adaptive fusion weight to reduce the impact of poor quality views.
In~\cite{guidolin2023hi} and~\cite{chen2022structural} constraints on human limbs length are exploited in the human pose estimation process in order to boost accuracy and robustness. They, however, require knowledge about subject's bones length or bone length statistics computed on the dataset of interest, making it difficult to apply such methods in unconstrained environments.

These methods typically do not generalize well to new scenarios and camera arrangements without extensive re-training and still struggle when dealing with strong occlusion. Therefore, they are not suitable to industrial multi-camera environments, which are prone to significant occlusions and in which collecting huge amounts of data for training may be impractical or even impossible.

\subsubsection{Absolute 3D Human Pose Estimation}
Recently, many 3D human pose estimation approaches have focused on single-view images or videos~\cite{gong2023diffpose, li2022mhformer, zhang2022mixste}. This is an ill-posed problem, affected by depth ambiguities and occlusions. Most works~\cite{kocabas2021pare, zhang20233d} mainly focus on root-relative human pose estimation, which provide human pose predictions with respect to a reference frame centered in the pelvis joint (root). Nevertheless, for real-world applications such as human-robot collaboration, it is crucial to accurately estimate the human body pose in the global 3D space~\cite{terreran2022skeleton, peng2023delving}. Pavlakos et al.~\cite{pavlakos2018learning} used a simple anthropometric approach, optimizing the absolute person distance based on 2D pixel positions and root-relative depth estimates from volumetric heatmaps, ensuring that the back-projected skeleton's bone lengths matched the average bone lengths from the training set. This method, however, has limited generalization power, as it requires pre-calculated statistics about human bone lengths.
Zhan \etal~\cite{zhan2022ray3d} use an intrinsic-parameter-invariant representation to convert the 2D keypoints from pixel space to 3D rays in a normalized 3D space; such rays are then fused from consecutive frames using temporal convolution. Sárándi \etal introduce MeTRAbs~\cite{sarandi2020metrabs, sarandi2023learning}, tackling scale and distance estimation of 3D poses while also addressing body occlusions. They train a fully-convolutional network to output location and depth heatmaps for 2D joints, providing accurate metric-scale depth information.
Still, when huge parts of the body are not visible, predicted scale and pose may be unreliable.

\subsubsection{Occlusion-Aware Human Pose Estimation}
Occlusion-aware human pose estimation has been primarily addressed by root-relative single-view methods.
To boost resilience to missing joints, synthetically occluded data can be exploited during training, as suggested in~\cite{sarandi2018robust}. Cheng \etal~\cite{cheng2019occlusion} make use of 2D confidence heatmaps of keypoints and introduce an optical-flow consistency constraint, to filter out the unreliable estimations. More recently, generative models have been successfully employed to retrieve occlusion-robust 3D pose, as in~\cite{zhang20233d}. Several single-view datasets representing scenarios with occlusions, such as 3DPW-Occ~\cite{zhang2020object} and 3DOH50K~\cite{zhang2020object} have also been proposed. Such works tackle root-relative 3D human pose estimation from monocular images and videos and mainly focus on everyday life scenarios.

Our work addresses the gaps in the current literature by proposing an occlusion-aware 3D human pose estimation method from multi-view images, targeting human-robot collaboration environments. Differently from current multi-view approaches, mostly relying on 2D feature triangulation, we aggregate single-view 3D predictions from absolute human pose estimation algorithms, to achieve robustness against strong occlusions.

\section{Method}
\label{sec:method}

The proposed method consists of three main parts: (i) single-view 3D pose prediction; (ii) multi-view 3D pose aggregation; (iii) 3D pose refinement based on reprojection error minimization.
In the first step, for each camera in the network, a candidate 3D pose is estimated by means of a monocular 3D human pose estimation algorithm, such as~\cite {sarandi2023learning, zhan2022ray3d}.
The 3D poses obtained from each view are then aggregated together by means of a dedicated multi-view fusion module, which aims to combine all the independent single-view predictions in a same 3D pose. Finally, the aggregated 3D pose is further refined in an optimization module based on constrained reprojection error minimization.

\subsection{Single-view 3D Pose Estimation}
\label{sec:method-sv}
As a first step, our approach takes as input multiple images captured simultaneously by $C$ cameras with known calibration parameters. Three-dimensional human poses are extracted from 2D images following a top-down approach: (i) a people detector finds all the possible patches containing a person, (ii) for each patch, a CNN predicts the 3D location of the human body joints with respect to the camera reference frame.
Our method assumes that all the cameras in the network are calibrated with respect to the same reference frame (e.g., robot base frame), so that the rotation $R_k$ and translation $t_k$ of each camera $k$ with respect to the reference frame are known.
For each view, the final output is a tuple of 3D poses, 2D bounding boxes and confidence scores.

\subsection{Multi-view Fusion Module}
\label{sec:method-mvf}
Monocular 3D human pose estimation methods typically provide accurate and plausible body joint predictions when considering relative joint positions.
However, they typically struggle in estimating the absolute distance of each body joints from the camera, especially in presence of large body occlusions.
Moreover, when portions of the human body are not visible, there exists an ambiguity between distance and size: small objects close to the camera may look similar to large objects farther away from the camera \cite{sarandi2020metrabs}, leading to incorrect 3D pose prediction in metric space.
In order to alleviate this problem, we propose to leverage information from multiple views, from the intuition that averaging the absolute positions of the single-view estimations should give a good approximation of the true location of the human body.

Consider a network of $C$ cameras and denote as $\{\hat{P}_{j,i}\}$ the set of 3D joints predicted from camera $i$. The $j^{th}$ joint of the fused skeleton can be expressed as the weighted average of the $j^{th}$ joint for all single-view detections:
%
\begin{equation}
\bar{P_{j}} = \frac{\sum_{i=1}^{C} w_i^j \hat{P}_{j,i}}{\sum_{i=1}^{C} w_i^j} ~.
\label{eq:weighted_sum}
\end{equation}
The use of weights $w_i$ allows to penalize the views framing the subject from a higher distance or the detections with lower confidence.
When the human body is partially or heavily occluded, in fact, the 3D pose obtained from monocular 3D pose estimation methods is not fully reliable: if some parts of the body are not visible, the pose estimation algorithm uses clues from the visible parts to estimate the positions of the missing joints. Significant occlusions reduce evidence for the obscured joints' positions, leading to incorrect final predictions compared to the true pose. For instance, if a person's legs are occluded by a table, the model lacks evidence to determine whether the person is standing with crossed, bent or straight legs.

To better incorporate all the available information, we propose the use of per-joint weights: if a 3D joint (\eg the right ankle) is incorrectly predicted in one view, the error given by projecting that joint over all the other views will be higher with respect to the reprojection error computed for that same joint correctly predicted from another view. 

In particular, given the set $\{\hat{P}_{j,i}\}$ of 3D joints predicted from camera $i$, a single joint $\hat{P}_{j,i}$ can be easily projected onto the image plane of a camera $k$ by means of the camera intrinsic parameters $K_k$ and extrinsic parameters $R_k$ and $t_k$. 
Averaging the reprojection error between the 2D  joints and the predictions on each frame $\{p_{j,i}\}$, allows to define the per-joint weight $w_{j,i}$ as follows:
\begin{equation}
w_{j,i} = \frac{1}{{e}_{j,i}}~, \qquad {e}_{j,i} = \frac{1}{C} \sum_{k=1}^{C} \left\| [{K_k}|\vec{0}][{R_k}|{t_k}] \hat{P}_{j,i} - {p}_{j,k} \right\|^2 ~.
\label{eq:reprojection_error}
\end{equation}

\subsection{Optimization Module}
The fusion module merges multi-view 3D predictions, leveraging multiple viewpoints to mitigate artifacts in joint positions caused by occluded body parts. However, when significant occlusions occur in multiple camera views, averaging predictions may not suffice, as estimated pose and camera distance can be slightly incorrect in all views. 
To address this issue, we formulate an optimization problem to refine the fused 3D pose $\bar{P_j}$, minimizing the per-joint reprojection error across all $C$ cameras available:
\begin{equation}
\underset{\bar{P_j}}{\mathrm{argmin}}\ \ \sum_{k=1}^{C} \sum_{j=1}^{J} \left\| [{K_k}|\vec{0}][{R_k}|{t_k}] \bar{P}_{j,i} - {p}_{j,k} \right\|^2 ~.
\label{eq:optimization_reprojection}
\end{equation}
Without additional constraints, the optimization process can change the position of all body joints; however, there are symmetries in the human body that can be exploited to guide optimization. In particular, we design a limb length symmetry constraint considering four pairs of bones: left and right upper arms, lower arms, upper legs and lower legs. For each pair we constrain the limb length to be the same, that is, we enforce left and right limbs to have consistent measures.
In particular, we compute the left (\emph{L}) and right (\emph{R}) 3D limb lengths as the euclidean distance $d(\hat{P}_{i}, \bar{P}_{j})$ between the corresponding left and right joints. If $u$ and $v$ are a pair of corresponding limb joints, the final cost for the symmetry constraint is given by: 
\begin{equation}
C_{sym} = \sum_{u=1}^J \sum_{v=1}^J \mathds{1}(u,v) \cdot \left\| d(\bar{P}_{u_L}, \bar{P}_{v_L}) - d(\bar{P}_{u_R}, \bar{P}_{v_R}) \right\|^2 ~,
\label{eq:optimization_symmetry}
\end{equation}

\begin{equation*}
\text{where} \ \ \mathds{1}(u,v) = 
\begin{cases}
    1 & \text{if } (u,v) \text{ is upper/lower arm, upper/lower leg} \\
    0 & \text{otherwise} ~.
\end{cases}
\label{eq:optimization_indice}
\end{equation*}
Note that, with respect to previous work \cite{pavlakos2018learning, chen2022structural, guidolin2023hi}, we do not require any prior information on human limbs length or pre-computed statistics on datasets' 3D annotations.
Finally, the optimization problem is defined as follows:
\begin{equation}
\underset{\bar{P_j}}{\mathrm{argmin}} \ \ \sum_{k=1}^{C} \sum_{j=1}^{J} \left\| [{K_k}|\vec{0}][{R_k}|{t_k}] \bar{P}_{j,i} - {p}_{j,k} \right\|^2 + C_{sym} ~.
\label{eq:optimization}
\end{equation}

\section{Experiments}
In order to evaluate the capabilities of our method to handle significant occlusions and clutter, following the work of~\cite{sarandi2018robust} and~\cite{zhang2020object}, we propose Human3.6M-Occluded, a novel benchmark derived from Human3.6M~\cite{ionescu2013human3}, adding synthetic occlusions on camera views. The robustness and reliability of our method in challenging human-robot environments is then inspected considering data sequences of human-robot collaboration tasks acquired in a real industrial workcell and in a laboratory environment. We finally conduct ablation studies to analyse the behaviour of our framework in presence of few occluded viewpoints and camera synchronization errors, typical in industrial multi-camera setups.

\subsection{Experimental Settings and Metrics}
\subsubsection{Human3.6M and Human3.6M-Occluded}
The Human3.6M dataset is currently one of the largest multi-view datasets providing 3D human pose annotations. It consists of 3.6 million frames captured by 4 synchronized 50\,Hz cameras. 3D and 2D keypoint ground truth is obtained by a marker-based motion capture system composed of 10 IR-cameras. 
The dataset includes 11 different subjects performing 15 daily actions, such as eating, walking, and giving directions. Each scene features only one subject.
Following the traditional protocol, subjects S9 and S11 are considered for evaluation. To provide fair comparisons with previous work~\cite{iskakov2019learnable}, \cite{ma2021transfusion}, sequences of subject S9 with incorrect 3D annotations (parts of `Greeting', `SittingDown' and `Waiting') were ignored for final results calculations. 

\begin{figure}[tb]
    \centering
    \begin{subfigure}[b]{0.22\textwidth}
        \centering
        \includegraphics[width=\textwidth]{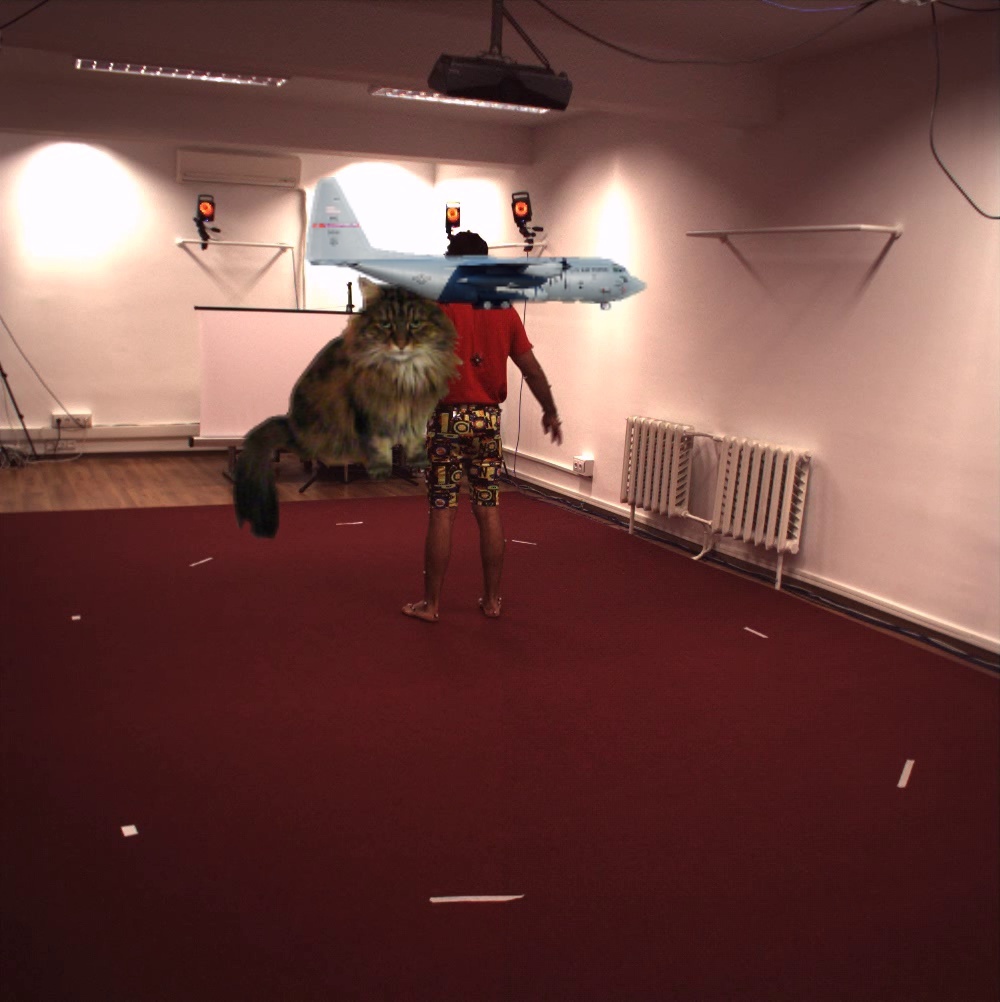}
        \caption{}
    \end{subfigure}
    \begin{subfigure}[b]{0.22\textwidth}
        \centering
        \includegraphics[width=\textwidth]{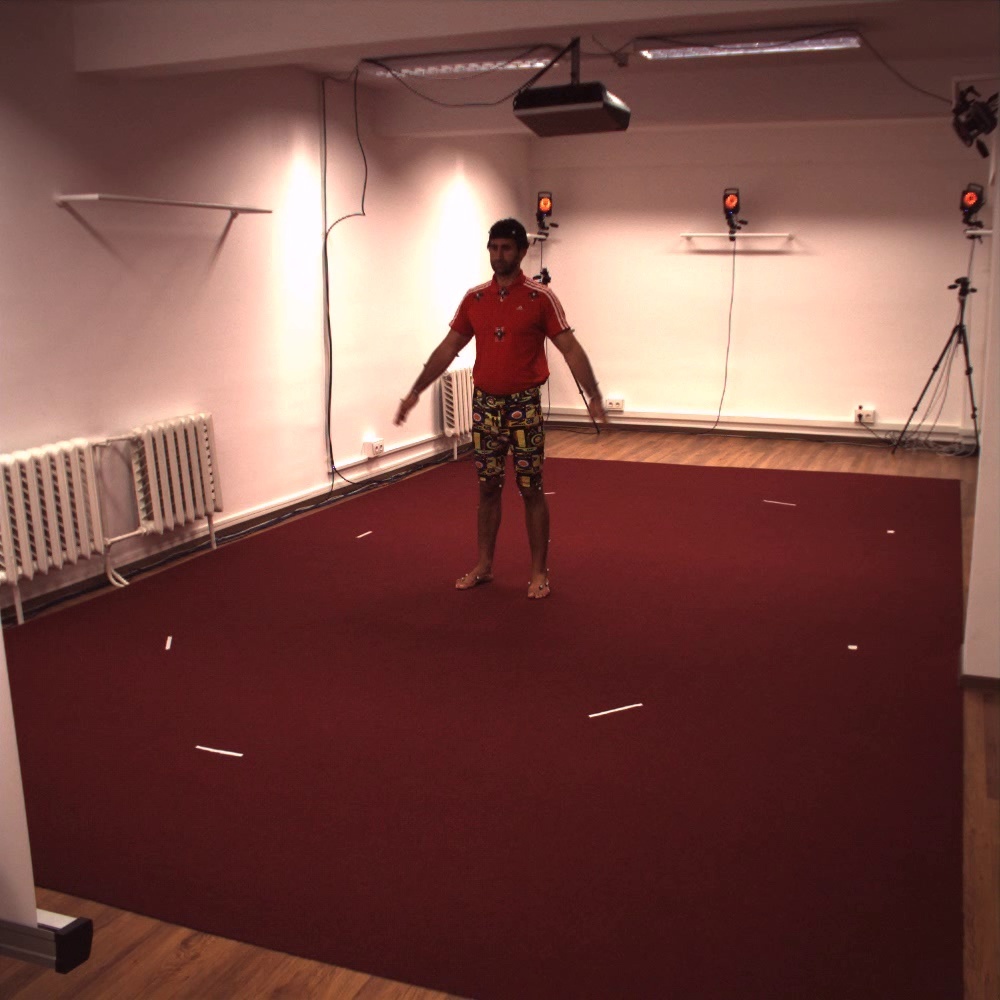}
        \caption{}
    \end{subfigure}
    \begin{subfigure}[b]{0.22\textwidth}
        \centering
        \includegraphics[width=\textwidth]{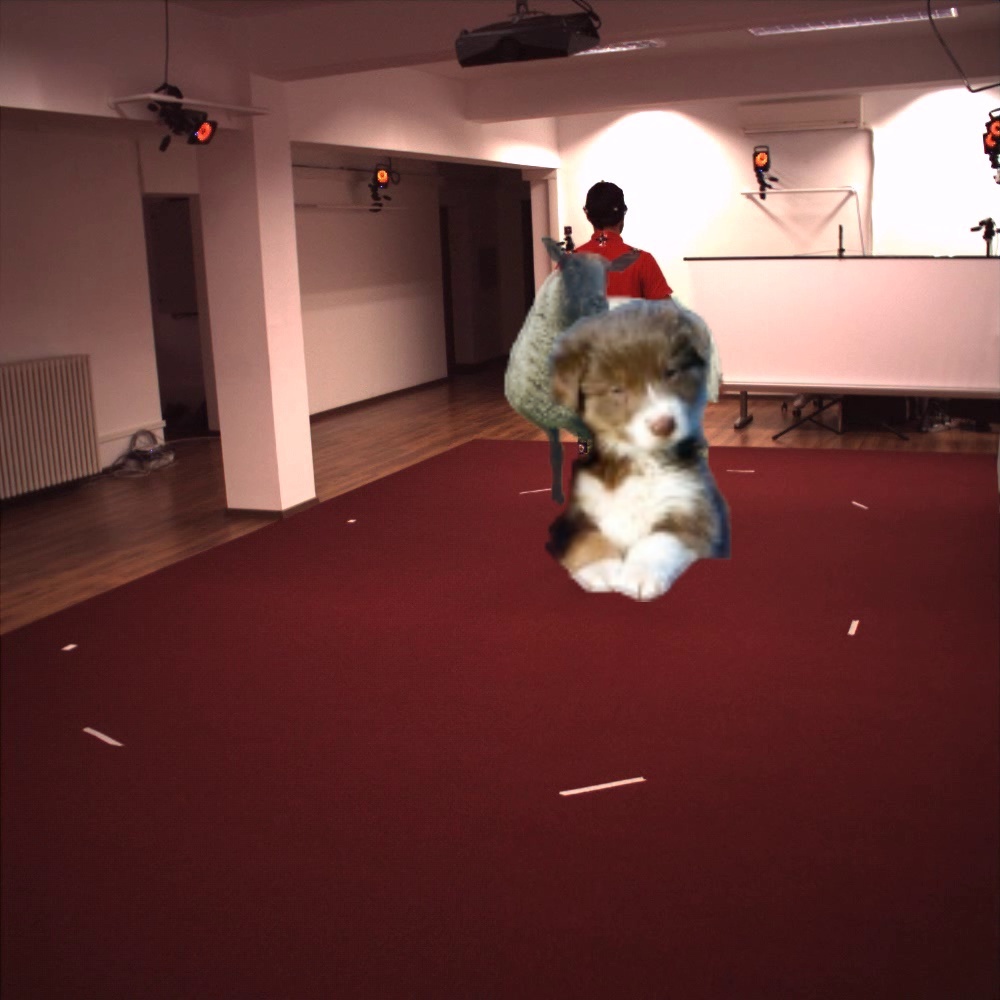}
        \caption{}
    \end{subfigure}
    \begin{subfigure}[b]{0.22\textwidth}
        \centering
        \includegraphics[width=\textwidth]{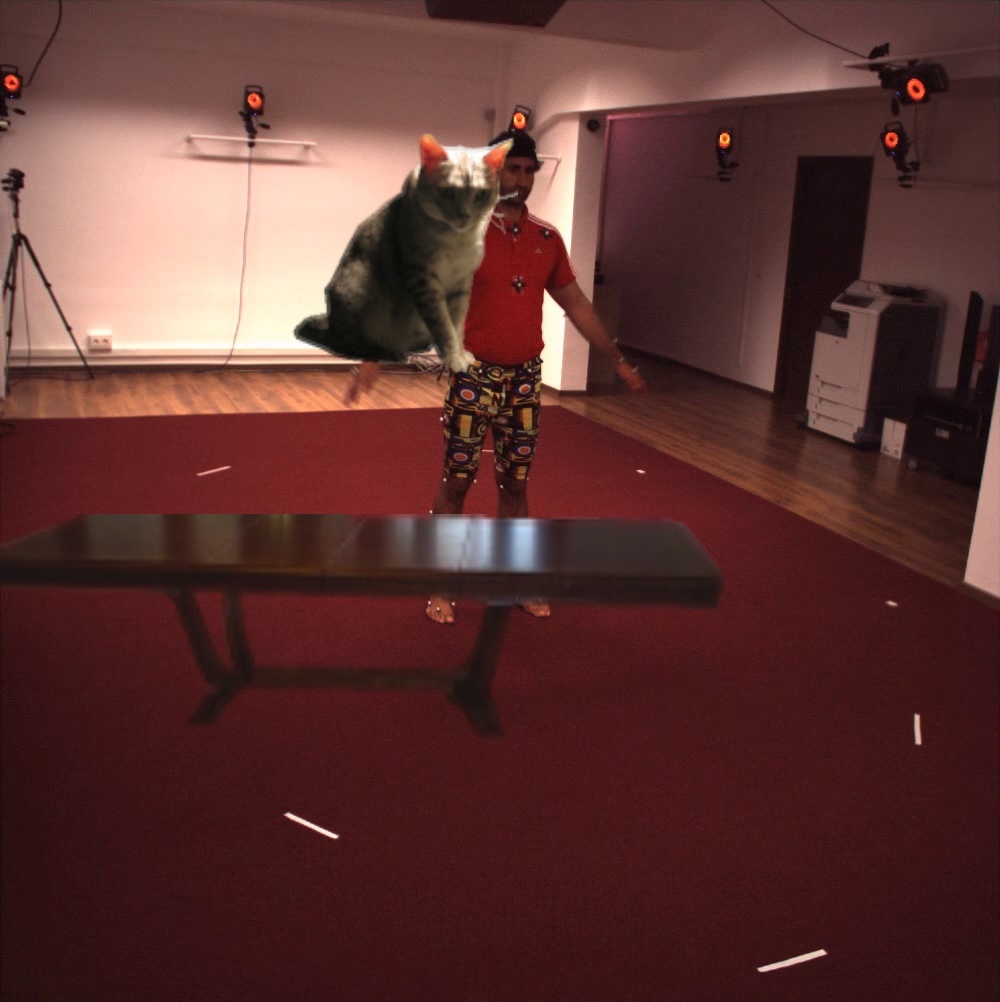}
        \caption{}
    \end{subfigure}
    \caption{Sample sequence from the Human3.6M-Occluded dataset, with occlusions on three of four views.}
\label{fig:h36m-occluded}
\end{figure}

Motivated by the lack of human body occlusions in Human3.6M and of public multi-view human pose estimation datasets including real occlusions in cluttered scenarios, we introduce the novel benchmark Human3.6M-Occluded. This is derived from the Human3.6M dataset by synthetically adding significant body occlusions, by taking objects and animals images from the Pascal VOC 2012 dataset~\cite{pascal-voc-2012} as occluders, as done in~\cite{sarandi2018robust}. Considering the four available camera views for each scene in the dataset, we partially cover the subject's body on three out of four views. This kind of scenario is very difficult to deal with, as each body joint is not guaranteed to be visible by at least two views. An occluded view includes two random objects overlapping the human bounding box. Object size and location inside the box are chosen at random.
Sample images from the Human3.6M-Occluded benchmark are shown in \Cref{fig:h36m-occluded}.

Differently from previous works presenting similar approaches, we make reproducible code available to generate the dataset, to encourage further research on occlusion-aware human pose estimation and to facilitate comparison with other methods. Code is available here \footnote{\url{https://github.com/laurabragagnolo/human3.6m-occluded}}.

\subsubsection{Human-Robot Workcell}
\begin{figure}[tb]
    \centering
    \begin{subfigure}[b]{0.4\textwidth}
        \centering
        \includegraphics[width=\textwidth]{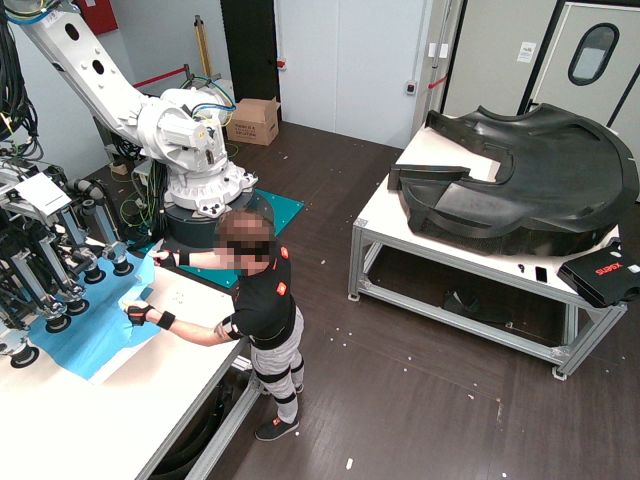}
        \caption{Workcell for collaborative draping}
        \label{fig:industrial}
    \end{subfigure}
    \begin{subfigure}[b]{0.4\textwidth}
        \centering
        \includegraphics[width=\textwidth]{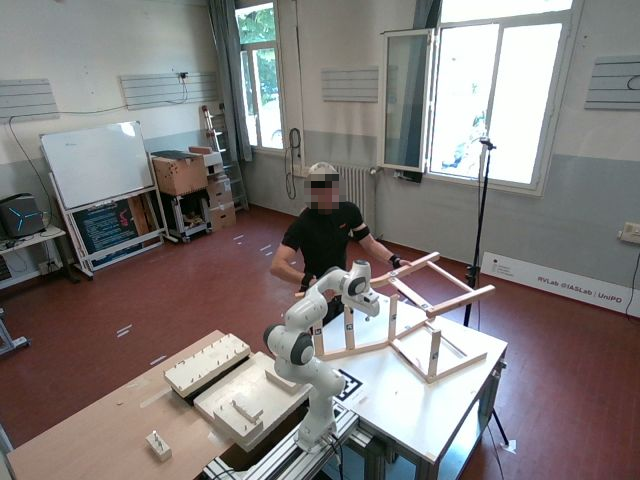}
        \caption{Workcell for collaborative assembly}
        \label{fig:laboratory}
    \end{subfigure}
    \caption{Sample images from the sequences acquired in real human-robot collaboration scenarios. a) Industrial workcell; b) Laboratory workcell.}
    \label{fig:workcell-images}
\end{figure}

To test the robustness of our human pose estimation method in real industrial environments, we conducted experimental validation on data acquired in two different human-robot collaboration scenarios: an industrial workcell developed as use case for the DrapeBot project~\cite{ghidoni2021smart, terreran2022smart}, and a laboratory collaborative workcell.
The former has been designed for collaborative transportation of flexible materials and covers an area of about 20\,m\(^2\). It was equipped with four Intel RealSense Depth D455 sensors surrounding an ABB industrial manipulator at an average distance of 4\,m from the base of the robot. 
The latter, equipped with four Intel RealSense Depth D455 cameras and a Franka Emika Panda collaborative manipulator, has been designed for the collaborative assembly of furniture, (\eg, wooden chair). 
The industrial workcell provides a very challenging testing scenario, with a lot of clutter in the scene and very tilted cameras. The laboratory workcell represents a simpler scenario, with less occluders.
In both settings cameras are not perfectly synchronized.
Sample images extracted from acquired sequences are shown in \Cref{fig:workcell-images}.

In all sequences collected, participants were equipped with the Xsens MVN Awinda motion capture system~\cite{schepers2018xsens} to record whole-body kinematics. The system consists of 17 wearable inertial measurement units (IMUs) placed all over the body, providing an accurate full-body 3D pose not affected by occlusions that has been used as a ground truth information. Since the 3D pose obtained from inertial sensors suffers from drifting errors in the absolute position over long period of time, we correct this effect using the information provided by the multi-camera pose tracking solution OpenPTrack~\cite{munaro2016openptrack}, by aligning the inertial-based 3D pose with the 3D position of the the neck joint detected by OpenPTrack, which has been empirically proven to be the most robust to occlusions among all body joints.

\subsubsection{Evaluation Metrics}
The 3D pose estimation accuracy is measured considering the Mean Per Joint Position Error (MPJPE), that is, the Euclidean distance between predicted 3D position and ground truth, for each joint.
We consider both the absolute and the root-relative MPJPE, that requires aligning the pelvis joints coordinates of the estimated and ground truth pose. While the absolute error takes into account the accuracy of the global localization of the prediction, the relative error focuses on the accuracy of the human keypoints configuration with respect to the pelvis joint, also known as the root joint.
The MPJPE is expressed in millimeters, where lower values represent higher accuracy.

\subsubsection{Implementation Details}
The single-view 3D predictions for multi-camera fusion are obtained using MeTRAbs~\cite{sarandi2023learning}. The considered model has been trained considering multiple skeleton conventions, \eg H36M~\cite{ionescu2013human3}, 3DHP~\cite{mehta2018single}, SMPL-X~\cite{pavlakos2019expressive}, and several 3D human pose estimation datasets, including Human3.6M~\cite{ionescu2013human3}, CMU Panoptic~\cite{joo2015panoptic} and TotalCapture~\cite{trumble2017total}.
For each of the experiments presented below, we do not perform any fine-tuning.

\subsection{Results on Human3.6M and Human3.6M-Occluded}
The results of the evaluation of our method and state-of-the-art multi-view human pose estimation algorithms on the Human3.6M dataset is reported in \Cref{tab:results-h36m-rel} and \Cref{tab:results-h36m-abs}. 
Considering the root-relative MPJPE (\Cref{tab:results-h36m-rel}), in scenarios with very few occlusions and in which the human body is well visible from all the four viewpoints, our method performs slightly worse with respect to current approaches. For multi-view methods, however, the absolute position error is much more interesting to look at, since it represents the quality of the 3D global localization of the human body prediction. Results for the absolute MPJPE are reported in \Cref{tab:results-h36m-abs}. Our framework surpasses most recent approaches and gives comparable results to strong baselines such as algebraic triangulation \cite{iskakov2019learnable}.
Looking the state of the art, it is worth noticing that our method and TransFusion \cite{ma2021transfusion} yield very similar results, while AdaFuse \cite{zhang2021adafuse} produces the strongest performance on this benchmark, considering both relative and absolute errors.

The loss of accuracy with respect to the state of the art is mainly due to the fact that competing approaches rely on heavy geometrical constraints, such as triangulation and epipolar lines, to perform 2D feature fusion. When the human body is fully visible and 2D features are very reliable, exploiting geometry gives extremely precise results that, however, poorly adapt to scenarios with occlusions, in which 2D joint predictions are not as reliable. Differently, our method uses information about the geometry of the scene, \ie, rotation and translation between cameras, for the sole 3D joint reprojection and then finds the optimal pose that minimizes reprojection errors on multiple views. This enforces looser geometrical constrains on the final pose, which lead to slightly worse results on the standard Human3.6M with respect to the other methods, but more robust performance in presence of occlusions.

\begin{table}[tb]
    \caption{The results of evaluation on the Human3.6M dataset. The table presents the MPJPE relative to pelvis for state-of-the-art multi-view human pose estimation methods.}
    \label{tab:results-h36m-rel}
    \resizebox{\columnwidth}{!}{%
        \begin{tabular}{@{}lcccccccccccccccc@{}}
        \toprule
        \multicolumn{17}{l}{MPJPE relative to pelvis, mm} \\ \midrule
        Methods & Dir. & Disc. & Eat & Greet & Phone & Photo & Pose & Purch. & Sit & SitD. & Smoke & Wait & WalkD. & Walk & WalkT. & \textbf{Avg} \\ \midrule
        RANSAC (as in \cite{zhang2021adafuse}) & 18.7 & 19.5 & 19.5 & 20.9 & 22.1 & 17.2 & 19.6 & 24.8 & 31.6 & 22.5 & 21.7 & 22.0 & \textbf{17.4} & 22.3 & 17.3 & 21.7 \\
        T-Iskakov \etal \cite{iskakov2019learnable} & 20.4 & 22.6 & 20.5 & 19.7 & 22.1 & 20.6 & 19.5 & 23.0 & 25.8 & 33.0 & 23.0 & 21.6 & 20.7 & 23.7 & 21.3 & 22.6 \\
        V-Iskakov \etal \cite{iskakov2019learnable} & 19.9 & 20.0 & 18.9 & 18.5 & 20.5 & 19.4 & 18.4 & 22.1 & \textbf{22.5} & 28.7 & 21.2 & 20.8 & 19.7 & 22.1 & 20.2 & 20.8 \\
        Zhang \etal \cite{zhang2021adafuse} & \textbf{17.0} & \textbf{18.1} & \textbf{17.3} & 21.1 & \textbf{19.1} & \textbf{16.4} & 18.1 & \textbf{19.2} & 26.3 & \textbf{19.4} & \textbf{19.1} & 21.0 & 16.8 & \textbf{19.6} & \textbf{16.9} & \textbf{19.4} \\
        Ma \etal \cite{ma2021transfusion} & 25.2 & 27.8 & 24.8 & 24.3 & 27.2 & 28.8 & 24.3 & 25.9 & 63.4 & 32.6 & 27.4 & 25.1 & 27.0 & 25.2 & 24.7 & 29.4 \\
        Wan \etal \cite{wan2023view} & 19.5 & 20.9 & 19.5 & \textbf{18.3} & 21.1 & 20.0 & \textbf{17.9} & 21.3 & 23.9 & 30.1 & 21.6 & \textbf{19.9} & 18.9 & 22.8 & 19.5 & 21.1 \\ \midrule
        \textbf{Ours} & 25.7 & 27.2 & 27.2 & 27.2 & 31.4 & 26.1 & 29.3 & 33.8 & 36.2 & 29.3 & 29.7 & 28.5 & 25.7 & 30.6 & 26.0 & 29.0 \\ \bottomrule
        \end{tabular}%
    }
\end{table}

\begin{table}[tb]
    \caption{The results of evaluation on the Human3.6M dataset. The table presents the absolute MPJPE for state-of-the-art multi-view human pose estimation methods.}
    \label{tab:results-h36m-abs}
    \resizebox{\columnwidth}{!}{%
        \begin{tabular}{@{}lcccccccccccccccc@{}}
        \toprule
        \multicolumn{17}{l}{MPJPE absolute, mm} \\ \midrule
        Methods & Dir. & Disc. & Eat & Greet & Phone & Photo & Pose & Purch. & Sit & SitD. & Smoke & Wait & WalkD. & Walk & WalkT. & \textbf{Avg} \\ \midrule
        RANSAC (as in \cite{zhang2021adafuse}) & 18.3 & 19.8 & 20.0 & 18.4 & 22.1 & 17.3 & 19.8 & 23.6 & 28.1 & 22.8 & 23.9 & 17.9 & 18.0 & 21.2 & 17.6 & 21.5 \\
        T-Iskakov \etal \cite{iskakov2019learnable} & 21.7 & 23.7 & 22.2 & 20.4 & 26.7 & 24.2 & 19.9 & 22.6 & 31.2 & 35.6 & 26.8 & 21.2 & 20.9 & 24.6 & 21.1 & 24.5 \\
        V-Iskakov \etal \cite{iskakov2019learnable} & 18.0 & \textbf{18.3} & \textbf{16.5} & \textbf{16.1} & \textbf{17.4} & 18.2 & \textbf{16.5} & \textbf{18.5} & \textbf{19.4} & 20.1 & \textbf{18.2} & 17.4 & \textbf{17.2} & \textbf{19.2} & \textbf{16.6} & \textbf{17.9} \\
        Zhang \etal \cite{zhang2021adafuse} & \textbf{17.4} & \textbf{18.3} & 17.8 & 17.8 & 19.4 & \textbf{16.8} & 18.5 & 19.2 & 22.1 & \textbf{19.4} & 21.4 & \textbf{16.7} & 17.4 & 19.6 & 17.4 & 19.4 \\
        Ma \etal \cite{ma2021transfusion} & 24.4 & 26.4 & 23.4 & 21.1 & 25.2 & 23.2 & 24.7 & 33.8 & 29.8 & 26.4 & 26.8 & 24.2 & 23.2 & 26.1 & 23.3 & 25.8 \\
        He \etal \cite{he2020epipolar} & 25.7 & 27.7 & 23.7 & 24.8 & 26.9 & 31.4 & 24.9 & 26.5 & 28.8 & 31.7 & 28.2 & 26.4 & 23.6 & 28.3 & 23.5 & 26.9 \\
        Moliner \etal \cite{moliner2024geometry} & - & - & - & - & - & - & - & - & - & - & - & - & - & - & - & 26.0 \\ \midrule
        \textbf{Ours} & 23.8 & 24.6	& 23.4 & 24.1 & 26.9 & 24.0 & 25.3 & 28.3 & 31.7 & 25.9 & 26.9	& 24.2	& 23.0	& 27.3	& 24.2	& 25.6 \\ \bottomrule
        \end{tabular}%
    }
\end{table}

To validate this, we now test our approach in presence of significant occlusions, on the proposed Human3.6M-Occluded benchmark.
For comparisons we select AdaFuse~\cite{zhang2021adafuse} and TransFusion~\cite{ma2021transfusion}, which have a specific focus on human pose estimation in unconstrained environments with occlusions and provide official codes and training weights on Human3.6M. We also compare with RANSAC (implementation from~\cite{zhang2021adafuse}) and the algebraic triangulation~\cite{iskakov2019learnable}, taken as baselines. 
For this experiment all the models are trained on the original Human3.6M dataset, without any fine-tuning on the Human3.6M-Occluded data.

Results for the relative and absolute MPJPE on this benchmark are shown in \Cref{tab:h36m-occ-rel} and \Cref{tab:h36m-occ-abs}. Our system provides the best performances with respect to both evaluation metrics, proving its superior capabilities in handling challenging scenes with occlusions with respect to competing methods relying on 2D features and strong geometrical constraints.

\begin{table}[tb]
    \caption{The results of evaluation on the proposed Human3.6M-Occluded benchmark. The table presents the MPJPE relative to pelvis for our method and state-of-the-art multi-view human pose estimation algorithms.}
    \label{tab:h36m-occ-rel}
    \resizebox{\columnwidth}{!}{%
        \begin{tabular}{@{}lcccccccccccccccc@{}}
        \toprule
        \multicolumn{17}{l}{MPJPE relative to pelvis, mm} \\ \midrule
        Methods & Dir. & Disc. & Eat & Greet & Phone & Photo & Pose & Purch. & Sit & SitD. & Smoke & Wait & WalkD. & Walk & WalkT. & \textbf{Avg} \\ \midrule
        RANSAC (as in \cite{zhang2021adafuse}) & 302.8 & 69.7 & 57.8 & 48.1 & 49.2 & 94.1 & 67.5 & 132.4 & 137.7 & 58.0 & 66.3 & 54.8 & 38.1 & 267.4 & 56.9 & 97.0 \\
        T-Iskakov \etal \cite{iskakov2019learnable} & 69.4 & 75.1 & 83.3 & 65.4 & 76.2 & 91.0 & 106.3 & 192.9 & 632.4 & 117.7 & 91.9 & 105.0 & 49.5 & 101.9 & 70.7 & 134.9 \\
        Zhang \etal \cite{zhang2021adafuse} & \textbf{32.5} & 41.3 & \textbf{33.9} & \textbf{35.0} & \textbf{37.7} & 38.0 & \textbf{36.4} & \textbf{50.5} & 83.0 & 47.4 & 47.9 & 38.7 & \textbf{28.9} & \textbf{40.3} & 37.6 & 43.5 \\
        Ma \etal \cite{ma2021transfusion} & 48.8 & 65.8 & 51.9 & 48.7 & 61.7 & 57.1 & 62.3 & 75.1 & 384.8 & 224.8 & 74.1 & 50.6 & 122.9 & 97.6 & 35.9 & 97.3 \\ \midrule
        \textbf{Ours} & 34.1 & \textbf{40.6} & 36.2 & 35.7 & 42.9 & \textbf{35.9} & 40.0 & 60.8 & \textbf{68.1} & \textbf{42.7} & \textbf{41.3} & \textbf{35.5} & 30.8 & 43.0 & \textbf{34.3} & \textbf{41.9} \\ \bottomrule
        \end{tabular}%
    }
\end{table}

\begin{table}[tb]
    \caption{The results of evaluation on the proposed Human3.6M-Occluded benchmark. The table presents the absolute MPJPE for our method and state-of-the-art multi-view human pose estimation algorithms.}
    \label{tab:h36m-occ-abs}
    \resizebox{\columnwidth}{!}{%
        \begin{tabular}{@{}lcccccccccccccccc@{}}
        \toprule
        \multicolumn{17}{l}{MPJPE absolute, mm} \\ \midrule
        Methods & Dir. & Disc. & Eat & Greet & Phone & Photo & Pose & Purch. & Sit & SitD. & Smoke & Wait & WalkD. & Walk & WalkT. & \textbf{Avg} \\ \midrule
        RANSAC (as in \cite{zhang2021adafuse}) & 305.5 & 61.3 & 51.9 & 47.2 & 46.6 & 130.0 & 62.2 & 117.5 & 117.5 & 53.7 & 65.5 & 55.0 & 35.2 & 60.2 & 51.6 & 80.7 \\
        T-Iskakov \etal \cite{iskakov2019learnable} & 67.1 & 71.3 & 80.4 & 63.4 & 74.8 & 88.5 & 100.9 & 188.7 & 611.2 & 111.7 & 92.2 & 110.6 & 47.1 & 99.7 & 69.8 & 127.4 \\
        Zhang \etal \cite{zhang2021adafuse} & \textbf{31.5} & 39.5 & \textbf{32.4} & 32.3 & \textbf{36.3} & 36.6 & \textbf{34.7} & \textbf{46.4} & 71.7 & 45.4 & 47.8 & 37.0 & 28.0 & \textbf{37.8} & 35.3 & 41.1 \\
        Ma \etal \cite{ma2021transfusion} & 48.1 & 65.2 & 50.9 & 47.9 & 60.8 & 56.1 & 61.6 & 74.1 & 384.2 & 224.2 & 73.4 & 49.9 & 122.3 & 96.5 & 35.8 & 96.5 \\ \midrule
        \textbf{Ours} & 32.0 & \textbf{36.9} & 32.9 & \textbf{31.5} & 37.9 & \textbf{32.9} & 36.2 & 53.7 & \textbf{64.9} & \textbf{37.5} & \textbf{38.9} & \textbf{32.8} & \textbf{27.8} & 38.5 & \textbf{31.4} & \textbf{37.8} \\ \bottomrule
        \end{tabular}%
    }
\end{table}

\subsection{Results on the Human-Robot Workcell Scenario}
We now inspect the robustness of our human pose estimation approach on real industrial environments. For this, we consider the data sequences collected in the industrial workcell and in our laboratory, presented earlier in this section.
Results are shown in \Cref{tab:workcell-rel} and \Cref{tab:workcell-abs}. Sequences acquired in the industrial workcell include the participation of two different subjects (performing two sequences each) and are numbered from 1 to 4; sequences collected in our laboratory include only one subject and are numbered from 5 to 7.

\begin{figure}[tb]
    \centering
    \includegraphics[width=\textwidth]{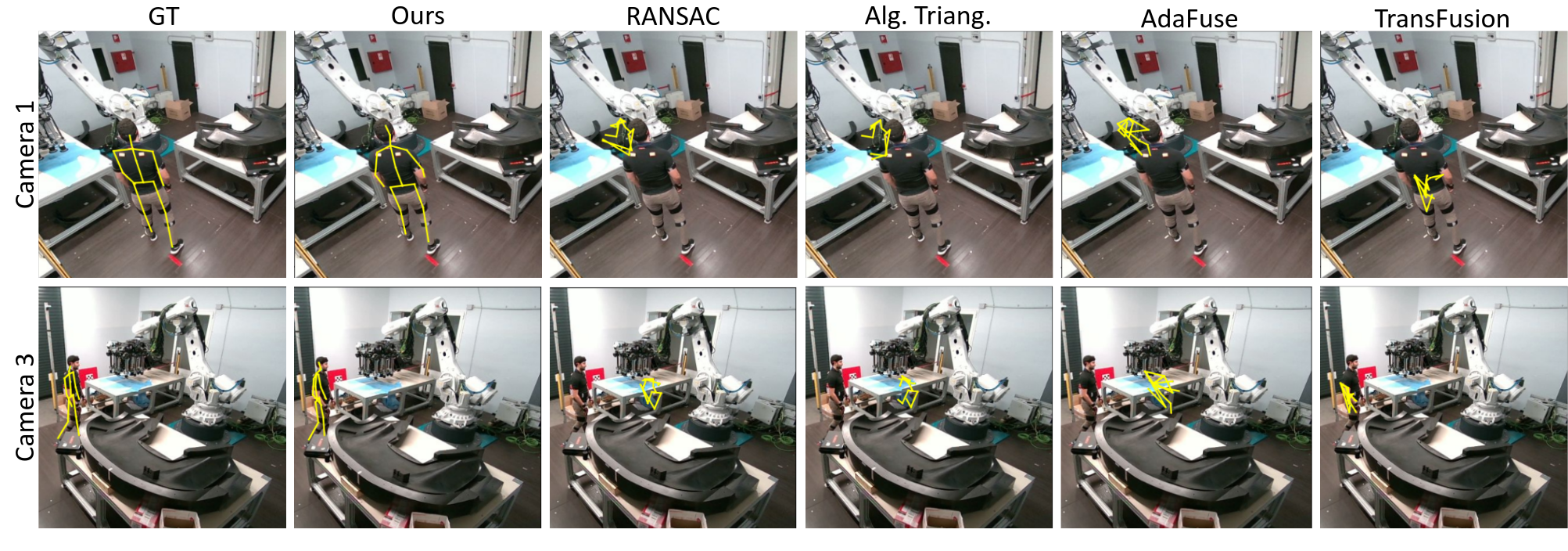}
    \caption{Qualitative comparison between our approach and the state of the art on the industrial workcell scenario.}
    \label{fig:industrial-workcell-results}
\end{figure}

\begin{figure}[tb]
    \centering
    \includegraphics[width=\textwidth]{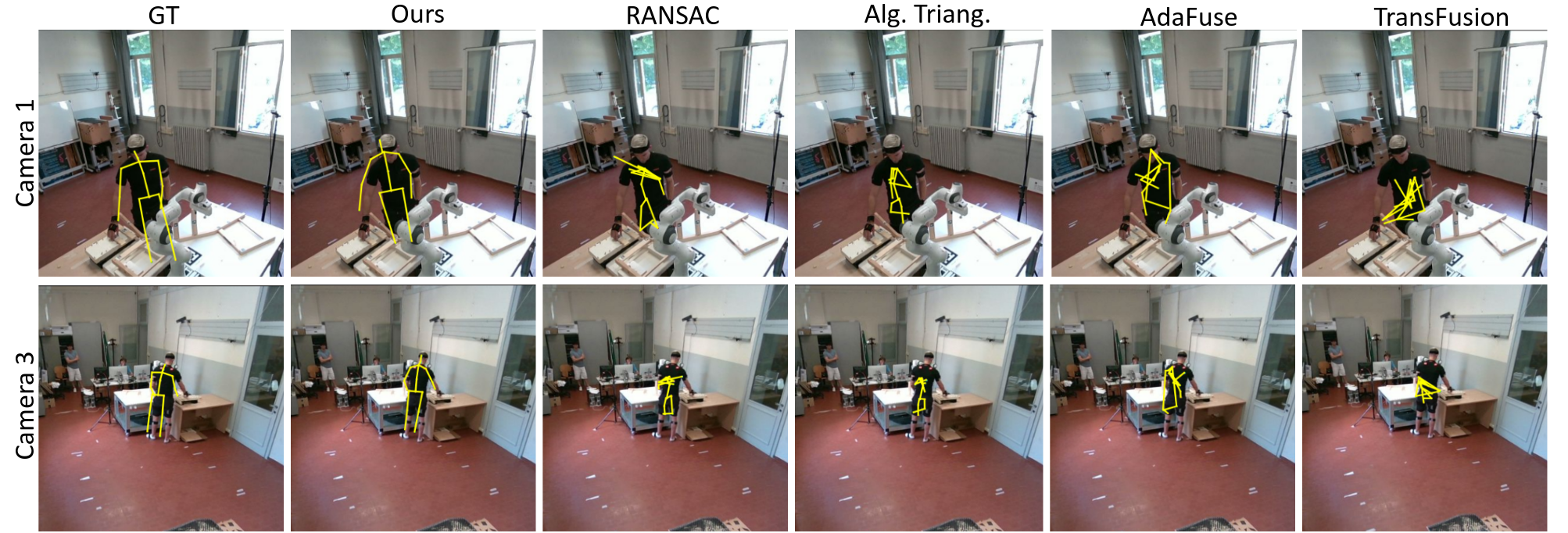}
    \caption{Qualitative results of our approach and state-of-the-art multi-view methods on the laboratory collaborative workcell.}
    \label{fig:lab-workcell-results}
\end{figure}

Our framework demonstrates a great advantage with respect to the other approaches, producing reliable predictions even on such challenging scenes, with an average absolute joint position error of about 14\,cm. On the other hand, the performance of the other methods dramatically decreases in the same context, leading to errors of several meters with respect to the ground truth reference, as shown in \Cref{fig:industrial-workcell-results} and \Cref{fig:lab-workcell-results}. 
Differently, our framework produces consistently better results on both scenarios, displaying excellent generalization power and adaptability to real multi-camera setups.

While results on Human3.6M-Occluded already confirmed the superior performance of our method in presence of significant occlusions, results on the real human-robot collaboration scenario really highlight the shortcomings of current methods and the true advantage of our framework. Nevertheless, the discrepancy that we observe between our results and those provided by the literature is due to conditions that characterize industrial multi-camera setups, such as camera synchronization errors and large occlusions on all viewpoints. These factors are not considered in public datasets~\cite{ionescu2013human3, joo2015panoptic}, and thus, not handled by most approaches in the literature.

\begin{table}[tb]
    \caption{The results of evaluation on data sequences collected in real human-robot workcells. The table presents the MPJPE relative to pelvis for our method and state-of-the-art multi-view human pose estimation algorithms.}
    \label{tab:workcell-rel}
    \centering
    \resizebox{0.7\columnwidth}{!}{%
        \begin{tabular}{@{}lcccccccc@{}}
        \toprule
        \multicolumn{9}{l}{MPJPE relative to pelvis, mm} \\ \midrule
        Methods & Seq 1 & Seq 2 & Seq 3 & Seq 4 & Seq 5 & Seq 6 & Seq 7 & \textbf{Avg} \\ \midrule
        RANSAC (as in \cite{zhang2021adafuse}) & 541.0 & 265.2 & 328.7 & 312.9 & 261.1 & 287.7 & 260.4 & 322.4 \\
        T-Iskakov \etal \cite{iskakov2019learnable} & 439.5 & 257.5 & 315.8 & 307.0 & 270.7 & 288.1 & 257.8 & 305.2 \\
        Zhang \etal \cite{zhang2021adafuse} & 580.7 & 370.0 & 402.1 & 447.0 & 347.2 & 344.9 & 321.7 & 401.9 \\
        Ma \etal \cite{ma2021transfusion} & 505.7 & 453.7 & 446.9 & 481.4 & 412.5 & 430.5 & 410.9 & 448.8 \\ \midrule
        \textbf{Ours} & \textbf{90.4} & \textbf{98.7} & \textbf{87.9} & \textbf{88.8} & \textbf{100.9} & \textbf{114.0} & \textbf{109.8} & \textbf{98.6} \\ \bottomrule
        \end{tabular}%
    }
\end{table}

\begin{table}[t]
    \caption{The table presents the absolute MPJPE on data collected in real human-robot workcells for our method and state-of-the-art multi-view human pose estimation algorithms.}
    \label{tab:workcell-abs}
    \centering
    \resizebox{0.7\columnwidth}{!}{%
        \begin{tabular}{@{}lcccccccc@{}}
        \toprule
        \multicolumn{9}{l}{MPJPE absolute, mm} \\ \midrule
        Methods & Seq 1 & Seq 2 & Seq 3 & Seq 4 & Seq 5 & Seq 6 & Seq 7 & \textbf{Avg} \\ \midrule
        RANSAC (as in \cite{zhang2021adafuse}) & 8981.5 & 7295.3 & 8109.3 & 7336.4 & 2800.5 & 2817.2 & 2813.5 & 5736.2 \\
        T-Iskakov \etal \cite{iskakov2019learnable} & 8784.7 & 7291.9 & 8112.4 & 7327.2 & 2796.3 & 2813.9 & 2808.4 & 5705.0 \\
        Zhang \etal \cite{zhang2021adafuse} & 9000.7 & 7276.1 & 8116.3 & 7308.6 & 2754.0 & 2765.7 & 2773.4 & 5713.5 \\
        Ma \etal \cite{ma2021transfusion} & 8907.5 & 7370.7 & 7296.0 & 8211.2 & 2788.6 & 2785.8 & 2786.8 & 5735.2 \\ \midrule
        \textbf{Ours} & \textbf{110.7} & \textbf{106.1} & \textbf{171.7} & \textbf{154.8} & \textbf{116.7} & \textbf{198.1} & \textbf{140.9} & \textbf{142.7} \\ \bottomrule
        \end{tabular}%
    }
\end{table}

\subsection{Ablation Studies}
In the remainder of this section we consider the Human3.6M and the Human3.6M-Occluded benchmarks to carry out ablation studies on different aspects typical of real industrial scenarios: (i) imperfect synchronization of cameras and (ii) heavy occlusions with few viewpoints available. 

\begin{figure}[tb]
    \centering
    \begin{minipage}[b]{0.49\textwidth}
        \centering
        \includegraphics[width=\textwidth]{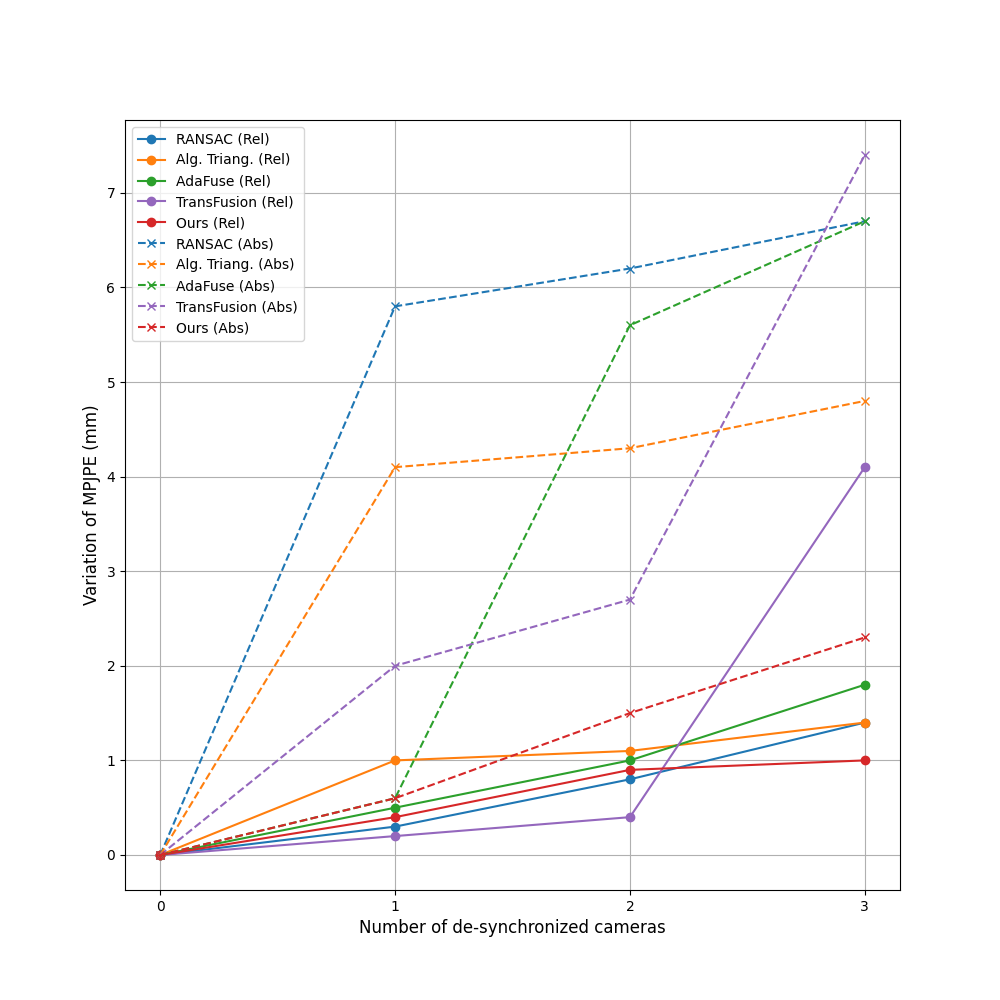}
        \caption{Variation of relative and absolute MPJPE with increasing number of de-synchronized cameras. Solid lines represent variation of the relative error, dashed lines represent variation of the absolute errors.\label{fig:abl-synch}}
    \end{minipage}
    \hfill
    \begin{minipage}[b]{0.49\textwidth}
        \centering
        \includegraphics[width=\textwidth]{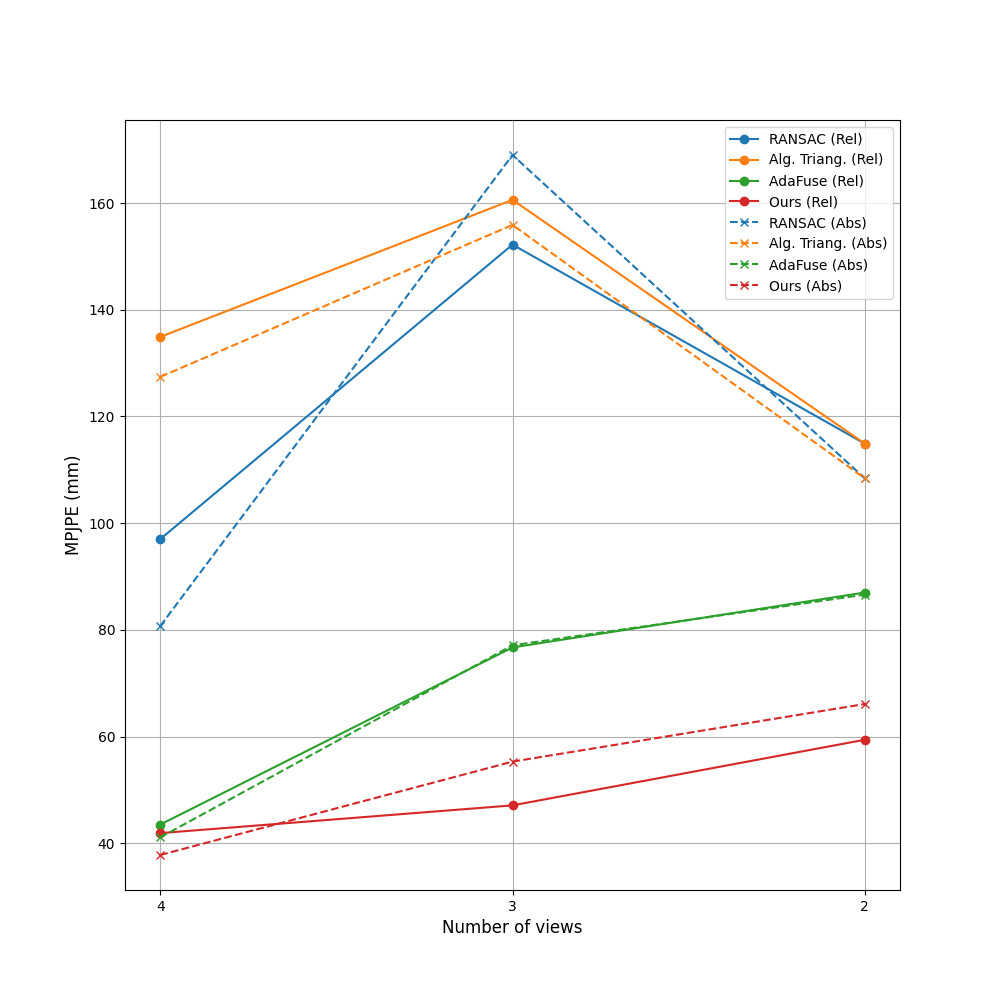}
        \caption{Relative and absolute MPJPE gradually removing the number of viewpoints, with occlusions. Solid lines represent variation of the relative error, dashed lines represent variation of the absolute errors.\label{fig:abl-views}}
    \end{minipage}
\end{figure}

\subsubsection{Camera Synchronization}
While public datasets are acquired in very constrained and controlled environments employing reliable camera synchronization mechanisms, such conditions are hardly met in industrial environments. To assess the advantages of our method, we now isolate the effect of slight errors in camera synchronization on multi-view human pose estimation methods.
Considering the Human3.6M dataset, we study the variation of the mean joint position error as we simulate synchronization errors on an increasing number of cameras.  
At first, three of the four cameras will give the correct frame corresponding to time $t$ while we simulate a synchronization error on the fourth one. The de-synchronized camera will give a frame corresponding to a wrong timestamp $t_e$, defined according to \Cref{eq:synch}.
\begin{equation}
\label{eq:synch}
    t_e = t + e \ \  \text{with} \ \ e \sim \text{Uniform}(\{-2, -1, +1, +2\})
\end{equation}
Then, we simulate a synchronization error on two of the four cameras and, finally, on three out of four cameras.
\Cref{fig:abl-synch} shows how the absolute and relative MPJPE vary over the three experiments. Our method proves to be the most robust to camera synchronization errors in terms of both relative and absolute MPJPE, which increase of only 1 and 2\,mm when three cameras out of four are unsynchronized.

\subsubsection{Number of Occluded Viewpoints}
To study the robustness of multi-view pose estimation approaches in presence of few occluded viewpoints, we consider Human3.6-Occluded and progressively reduce the number of available cameras to three and to two.
It is worth noticing that in Human3.6M-Occluded, for each scene, three out of four views contain occlusions and the occluded cameras are selected at random. This means that when removing a camera, it can happen that for some scenes all views are occluded.

\Cref{fig:abl-views} shows the relative and absolute MPJPE with reduced number of viewpoints for the compared methods. For a fair comparison, we do not consider TransFusion~\cite{ma2021transfusion} for this ablation study, as this approach requires to be re-trained when the number of available cameras change.
One can easily notice that our framework is the most robust to significant occlusions with few viewpoints, yielding a relative and absolute joint position error of about 6\,cm when tested on scenes with only two cameras and occlusions on at least one view. 

\section{Conclusion}
In this work, we proposed a novel multi-view 3D pose fusion approach for robust and reliable 3D human pose estimation in human-robot collaboration scenarios, generally characterized by significant occlusions and limited camera viewpoints. While most methods rely on 2D keypoints triangulation, we aggregate single-view 3D joint predictions, proposing a per-joint weighting depending on the mean reprojection error on all views, to mitigate the effect of incorrect predictions. The fused 3D pose is further refined via reprojection error optimization, introducing limb length symmetry constraints to enhance estimation accuracy.

Our framework was evaluated on challenging scenarios with sever occlusions, considering data sequences from real human-robot collaboration workcells and the proposed benchmark Human3.6M-Occluded. Our method outperforms state-of-the-art techniques on heavy occlusions and exhibits strong generalization capabilities to unseen environments, making it a suitable solution for 3D human pose estimation in challenging human-robot collaboration settings.

\section*{Acknowledgements}
The research leading to these results has received funding from the European Union’s Horizon 2020 research and innovation programme under grant agreement No. 101006732 (DrapeBot).

%
%
\bibliographystyle{splncs04}
\bibliography{main}
\end{document}